\documentclass[conference]{IEEEtran}
\IEEEoverridecommandlockouts
\usepackage{cite}
\usepackage{amsmath,amssymb,amsfonts}
\usepackage{algorithmic}
\usepackage{textcomp}
\usepackage{float}
\usepackage{subfig} 
\usepackage{placeins}

\usepackage{subcaption}

\usepackage{url}
\usepackage{hyperref}
\hypersetup{breaklinks=true}
\usepackage{subfig} % Use the subfig package

\usepackage{graphicx}

\usepackage{booktabs} % for professional tables

\usepackage{microtype}

\usepackage{multirow}
\usepackage[table,xcdraw]{xcolor}
\usepackage[normalem]{ulem}
\useunder{\uline}{\ul}{}
\usepackage{comment}
\usepackage{amsmath,amssymb} % define this before the line numbering.
\usepackage{color}
\usepackage{fancyhdr}

\fancypagestyle{firstpage}{
  \fancyhf{} % Clear all header and footer fields
  \fancyhead[C]{2024 International Conference on Machine Learning and Applications (ICMLA)} % Center header
  \fancyfoot[L]{1946-0759/24/\$31.00 \copyright 2024 IEEE \\ DOI 10.1109/ICMLA61862.2024.00082} % Left footer
  \fancyfoot[C]{\thepage} % Center footer (page number)
  \renewcommand{\headrulewidth}{0pt} % Remove header rule
  \renewcommand{\footrulewidth}{0pt} % Remove footer rule (optional, as there’s typically no footer rule by default)
}

\def\BibTeX{{\rm B\kern-.05em{\sc i\kern-.025em b}\kern-.08em
    T\kern-.1667em\lower.7ex\hbox{E}\kern-.125emX}}

\begin{document}

\pagestyle{fancy}
\fancyhf{} % Clear all header and footer fields
\fancyfoot[C]{\thepage} % Page number at bottom center
\renewcommand{\headrulewidth}{0pt} % Remove header rule
\renewcommand{\footrulewidth}{0pt} % Remove footer rule (optional)

\setcounter{page}{560}

%\title{A Framework for Investigating the Generalizability of Self-Supervised Learning}
%\title{Self-Supervised Learning for Recognizing Wild Animals from Noisy ``Selfies''}
%\title{Investigating Self-Supervised Learning on Real-World Noisy Datasets}
%\title{Leveraging Language Models for Analysis of Experiential Time-Series Data in Education}

\title{Leveraging Language Models for Analyzing Longitudinal Experiential Data in Education}
%\title{Leveraging Language Models for Analyzing Experiential Time-Series Data in Education}

%\title{Leveraging Language Models for Forecasting Educational Outcomes Using Experiential Time-Series Data}

%\title{Conference Paper Title*\\

\author{\IEEEauthorblockN{Ahatsham Hayat}
\IEEEauthorblockA{\textit{Electrical and Computer Engineering} \\
\textit{University of Nebraska-Lincoln}\\
aahatsham2@huskers.unl.edu}
\and
\IEEEauthorblockN{Bilal Khan}
\IEEEauthorblockA{\textit{Computer Science and Engineering} \\
\textit{Lehigh University}\\
bik221@lehigh.edu}
\and
\IEEEauthorblockN{Mohammad Rashedul Hasan}
\IEEEauthorblockA{\textit{Electrical and Computer Engineering} \\
\textit{University of Nebraska-Lincoln}\\
hasan@unl.edu}
%\and
%\and
%\and
}

\maketitle

\thispagestyle{firstpage}

\begin{abstract}

We propose a novel approach to leveraging pre-trained language models (LMs) for early forecasting of academic trajectories in STEM students using high-dimensional longitudinal experiential data. This data, which captures students' study-related activities, behaviors, and psychological states, offers valuable insights for forecasting-based interventions. Key challenges in handling such data include high rates of missing values, limited dataset size due to costly data collection, and complex temporal variability across modalities. Our approach addresses these issues through a comprehensive data enrichment process, integrating strategies for managing missing values, augmenting data, and embedding task-specific instructions and contextual cues to enhance the models' capacity for learning temporal patterns. Through extensive experiments on a curated student learning dataset, we evaluate both encoder-decoder and decoder-only LMs. While our findings show that LMs effectively integrate data across modalities and exhibit resilience to missing data, they primarily rely on high-level statistical patterns rather than demonstrating a deeper understanding of temporal dynamics. Furthermore, their ability to interpret explicit temporal information remains limited. This work advances educational data science by highlighting both the potential and limitations of LMs in modeling student trajectories for early intervention based on longitudinal experiential data.

\end{abstract}

\begin{IEEEkeywords}
Language Model, Time-series Data, Experiential Data, Missing Data, Data Augmentation, STEM Education
\end{IEEEkeywords}

\vspace{-2.00mm}

\section{Introduction}
\label{intro}

``We apprehend more than we comprehend.''
\vspace{-2.00mm}
 \begin{flushright} \textit{Michel Serres, The Parasite (1982)}\end{flushright}
                                            
\vspace{0.50mm}
%Experience plays a central role in the human learning process~\cite{kolb_1984_experiential}. Gaining insights into learners' experiences can significantly enhance learning outcomes~\cite{palmer_2010, petrina_2018_blog}. This is often achieved by analyzing \textbf{experiential data}, which encompasses learners' perceptions and interactions within their learning activities~\cite{andrews_2009}. Such data sheds light on the subjective and qualitative dimensions of learning, including emotions, perceptions, opinions, and values linked to their educational engagements. Moreover, it captures the extralinguistic aspects of a learner's journey, uncovering the strengths, weaknesses, opportunities, and challenges inherent in their experiences. This, in turn, offers pathways for enhancing educational experiences. For instance, understanding students' academic experiences through such data can lead to just-in-time intervention strategies, potentially predicting their cognitive performance well before the semester's end~\cite{wang_smartgpa_2016, wang_studentlife_2014, li_student_2020, xu_application_2022}.

Experience plays a central role in the human learning process~\cite{kolb_1984_experiential}. Gaining insights into learners' experiences can significantly enhance learning outcomes~\cite{palmer_2010, petrina_2018_blog}. This is often achieved by analyzing \textbf{longitudinal experiential data}, which involves systematically aggregating real-time observations from individuals over time through methods such as self-report surveys. This data encompasses learners' perceptions and interactions within their learning activities~\cite{andrews_2009}. Such rich, time-varying data on human experience provides insights into the subjective and qualitative dimensions of learning, including emotions, perceptions, opinions, and values related to their educational engagements. Additionally, it captures extralinguistic aspects of a learner's journey, uncovering their strengths, weaknesses, opportunities, and challenges. This, in turn, offers pathways for enhancing educational experiences. For example, understanding students' academic experiences through this data can inform just-in-time intervention strategies, potentially predicting cognitive performance well before the end of the semester~\cite{wang_smartgpa_2016, wang_studentlife_2014, li_student_2020, xu_application_2022}.

Leveraging longitudinal experiential data to forecast academic performance with artificial intelligence (AI), particularly through deep learning (DL) techniques, presents notable challenges. Firstly, since experiential data primarily consists of qualitative text, advanced DL-based natural language processing (NLP) techniques are required to effectively interpret and utilize this non-numeric information. Secondly, the inherent temporal dynamics within longitudinal experiential data, characterized by repeated measurements of experiential attributes, transform it into a complex time-series dataset. This complexity necessitates the development of innovative DL-based approaches for time-series forecasting that are specifically designed to accommodate and learn from qualitative inputs. 
%These challenges underscore the need for novel approaches in both NLP and time-series analysis within the context of DL to accurately predict academic outcomes from experiential data.

Recently, Transformer-based~\cite{vaswani_attention_2017} pre-trained language models (LMs)~\cite{touvron_llama_2023, openai_gpt-4_2023} have revolutionized various AI domains, including time-series forecasting~\cite{jin_2023_LLM_survey, zhou_2023_OFA, gruver_2023_llm, jin_2023_timellm, chang_2023_llm4ts, cao_2023_tempo, sun_2023_test}. However, the suitability of these models for creating intervention systems based on qualitative longitudinal experiential data remains under-explored. Most existing methods are tailored to numeric, non-experiential longitudinal data, indicating a gap in the application of pre-trained LMs to the nuanced and text-rich domain of experiential data. 
%This gap highlights the need for research into adapting these advanced LMs to effectively process and forecast outcomes from qualitative experiential data streams.

%brown_language_2020

%chowdhery_palm_2022

In this research, we explore the extent to which pre-trained LMs can effectively interpret and utilize longitudinal experiential data within the context of student learning, specifically in STEM (science, technology, engineering, and mathematics) education. Addressing the lack of suitable high-dimensional experiential datasets for such research, we have compiled a unique \textbf{78-dimensional dataset} that encompasses a holistic view of college students' academic journey over a semester.
%\footnote{The dataset will be available upon paper acceptance.}

This dataset is divided into three key components: (i) The non-cognitive component, which includes 28 dimensions of repeated qualitative measurements, captures attributes such as student motivation and engagement, offering insights into students' perceptions of their academic experiences; (ii) The cognitive component, consisting of 41 quantitative measures, encompasses students' formative and summative assessment scores; and (iii) The background factors component, featuring 9 dimensions of qualitative data, provides static information on students' academic meta-information and socioeconomic status.

%non-experiential segment consists of (i) a 41-dimensional quantitative analysis of cognitive attributes, including students' formative and summative assessment scores, and (ii) a 9-dimensional qualitative assessment of static data regarding students' background factors, such as academic meta-information and socioeconomic status.

Crucially, both the non-cognitive and cognitive data components are structured as \textbf{time-series}. Utilizing this comprehensive, high-dimensional dataset, our research aims to determine if pre-trained LMs can effectively decipher experiential cues for early forecasting of students' end-of-semester cognitive performance. Specifically, we examine the models' ability to learn and integrate the complex correlations between non-cognitive and cognitive data elements, and to account for their temporal variations. This exploration is pivotal in understanding how advanced LMs can adapt to the nuanced domain of educational data, which encompasses both qualitative and quantitative aspects, thereby facilitating the development of effective just-in-time interventions.
%\vspace{-1.00mm}

The nature of our longitudinal experiential student data presents \textbf{unique challenges}, distinctly setting it apart from the time-series numeric data typically employed by state-of-the-art time-series LM-based methods~\cite{jin_2023_LLM_survey}. Our dataset's distinctive characteristics include: (i) A hybrid structure combining static background features with time-variant cognitive and non-cognitive elements; (ii) The inclusion of non-numeric, experiential measurements, (iii) The forecasted variable is text-based assessments of future summative performance; (iv) A significant proportion of missing values within the non-cognitive data, complicating the learning of correlations with the time-series cognitive components (discussed in Section \ref{dataset}); (v) A lack of temporal alignment between non-cognitive and cognitive modalities, with respective cross-modality features often recorded on different days; (vi) A relatively small dataset size (N=48), posing challenges for effective LM-based transfer learning due to the high cost of collecting comprehensive longitudinal data.

To address these challenges and harness the general knowledge and reasoning capabilities of pre-trained LMs~\cite{raffel_exploring_2020, roberts_how_2020, wei_chain--thought_2023, bhatia_tart_2023}, we develop a \textbf{data enrichment} method. This approach enables fine-tuning pre-trained LMs for early performance forecasting by reframing cognitive performance forecasting as a language generation problem. Our data enrichment method involves: (i) Handling of missingness in student experiential data; (ii) Augmentation of text sequence data to counter the limitations posed by the dataset's small size; (iii) Inclusion of explicit task instructions and contextual cues to guide LMs in understanding the task, recognizing temporal orders, and applying domain-specific knowledge, thereby addressing learning challenges across different data dimensions.

%chowdhery_palm_2022

In our comprehensive empirical study, we evaluate two types of pre-trained LMs – decoder-only and encoder-decoder models – to systematically examine their capability in early forecasting of summative cognitive performance by utilizing experiential data. We address the following pivotal research questions (RQs):

\begin{itemize}
\item \textbf{[RQ1]}: To what extent can LMs accurately forecast outcomes based solely on longitudinal experiential data?
\item \textbf{[RQ2]}: How effectively do LMs capture and leverage correlations across the non-cognitive, cognitive, and background modalities within academic experiential data for precise early forecasting?
\item \textbf{[RQ3]}: What is the extent of LMs' ability to interpret and use temporal variations within the dataset for forecasting purposes?

\item \textbf{[RQ4]}: How can we effectively address missingness in experiential datasets by leveraging pre-trained LMs?

%\item \textbf{[RQ4]}: How do LMs utilize contextually relevant descriptors to compensate for missing values in the data?
\end{itemize}
%Our key contributions include the creation of a multi-dimensional experiential dataset focused on STEM education, the development of a novel data enrichment method tailored for pre-trained LMs, and a set of extensive empirical studies that shed light on the learning behaviors of decoder-only and encoder-decoder LMs. A significant observation from our studies is the impressive capability of LMs to assimilate experiential data and effectively discern correlations across experiential and non-experiential modalities. However, while LMs are adept at identifying temporal patterns in time-series data, they tend to rely predominantly on surface-level statistical relationships. For instance, our findings indicate that LMs do not effectively utilize explicit temporal tags embedded in the dataset to learn time-series patterns. This suggests that within the scope of experiential data, the proficiency of LMs is more aligned with uncovering complex statistical correlations, rather than comprehending the deeper semantic context of the data, which is likely to impact the development of just-in-time intervention systems in education by utilizing LMs.

Our key contributions include the creation of a multi-dimensional longitudinal experiential dataset focused on STEM education, the development of a novel data enrichment method tailored for pre-trained LMs, and a set of extensive empirical studies that shed light on the learning behaviors of decoder-only and encoder-decoder LMs. A significant observation from our studies is the impressive capability of LMs to assimilate experiential data and effectively discern correlations across the modalities of experiential data. However, while LMs are adept at identifying temporal patterns in time-series data, they tend to rely predominantly on surface-level statistical relationships. For instance, our findings indicate that LMs do not effectively utilize explicit temporal tags embedded in the dataset to learn time-series patterns. This limitation underscores that within the scope of longitudinal experiential data, LMs excel more at uncovering complex statistical correlations than at comprehending deeper semantic contexts, a crucial aspect for the advancement of just-in-time intervention systems in education using LMs.

\section{Dataset Development \& Enrichment}
\label{dataset}

\vspace{-3.00mm}

\begin{figure}[htb!]
\centering
\includegraphics[width=1.0\columnwidth]{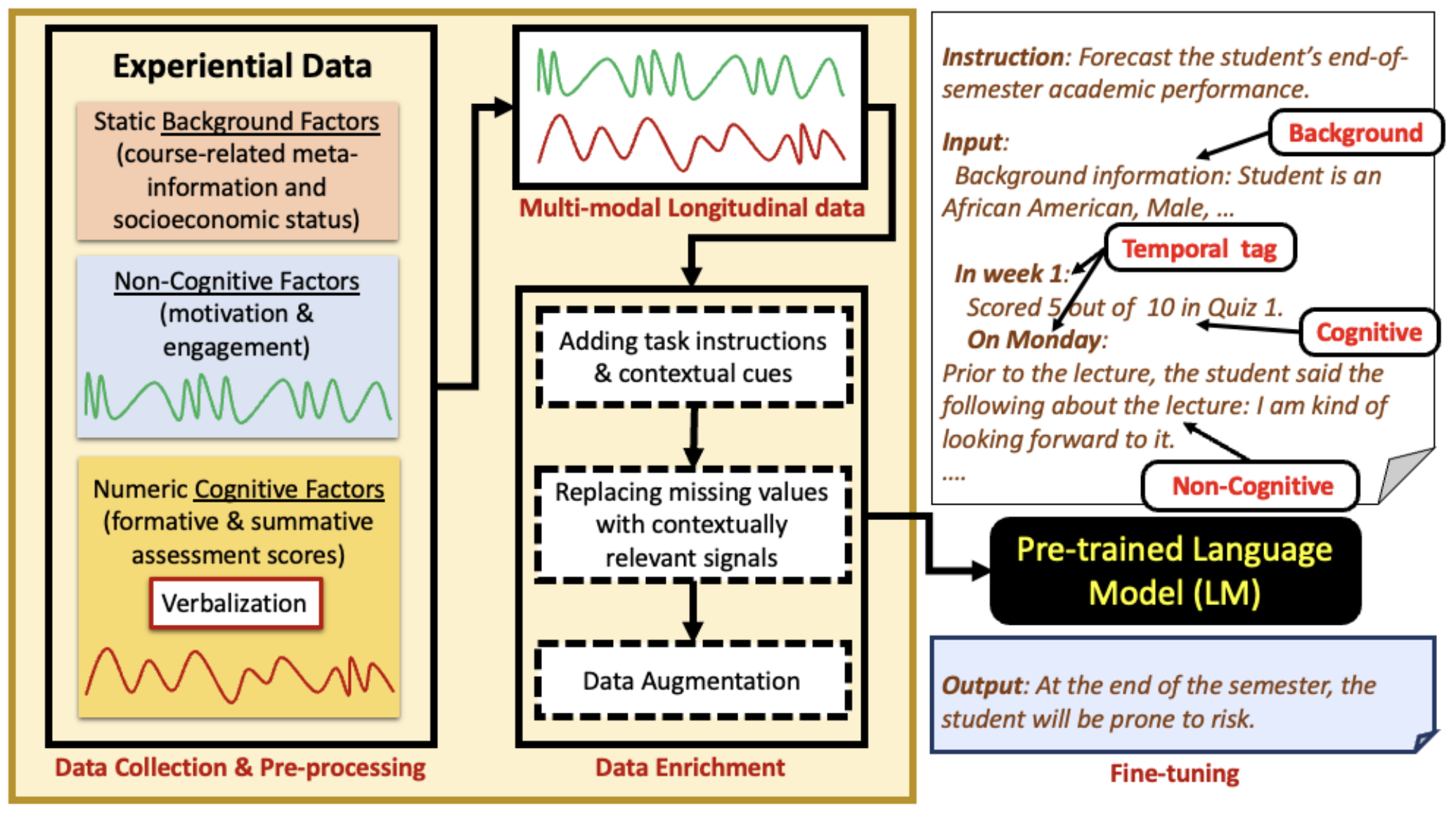}
%\vspace{-2.00mm}
\caption{Overview of the longitudinal experiential dataset development process that is amenable to adapting LMs.}
\vspace{-2.00mm}
\label{fig:pipeline}
\end{figure}

To enable pre-trained LMs to gain a nuanced understanding of students' academic experiences, we compiled a comprehensive dataset that explores the interplay between experiential modalities in student learning. Figure \ref{fig:pipeline} provides an overview of our dataset development process, including data collection, pre-processing, and a method for data enrichment. The transformed data is utilized to adapt pre-trained LMs for early forecasting of students' end-of-semester cognitive performance based on the first 4 weeks of data. Performance is categorized into four types, aligned with major letter grade thresholds: at-risk (grade C or below), prone-to-risk (above C but below B), average (above B but below A), and outstanding (grade A or above).

\subsection{Data Collection}
We gathered data from 48 first-year college students enrolled in an introductory programming course at a public university in the U.S. The dataset, comprising 78-dimensional data, captures three modalities of students' academic experiential trajectories, as described below.

%We gathered data from 48 first-year college students enrolled in an introductory programming course (CS1) at a public university in the USA. The study received approval from the University of XXX's Institutional Review Board (IRB \#: XXX). The dataset, comprising 78-dimensional data, captures three modalities of students' academic experiential trajectories, as described below.

\vspace{2.00mm}
\textbf{Background Data (9-dimensional)}: At the semester's start, we collected essential 9-dimensional background data via a Qualtrics-based web survey. This data includes course-related meta-information (class standing and major) and socioeconomic factors (gender, race, international or native student status, parents' education background, highest education level of a single parent, highest education level of another parent, family yearly income, science identity, and reflected science identity). These factors are included to examine the \textbf{impact of personalization} on LMs, hypothesizing that these attributes correlate with students' academic trajectories and future course performance~\cite{bandura_social_2001}, serving as valuable priors for LMs to recognize individualized patterns in academic progression.

\vspace{2.00mm}
\textbf{Cognitive Data (41-dimensional)}: This data includes 41-dimensional cognitive data from students' assessment scores (formative and summative) throughout the 16-week semester, sourced from the course's learning management system.

\vspace{2.00mm}
\textbf{Non-Cognitive Data (28-dimensional)}: The 28-dimensional non-cognitive data comprises repeated measures of students' motivation (intrinsic and extrinsic) and engagement (behavioral, emotional, and cognitive) factors across the semester. This selection aims to capture students' evolving study-related behaviors, with research indicating a strong correlation between these non-cognitive factors and students' learning outcomes~\cite{fogg_behavior_2009, fredricks_eight_2014}. This correlation is essential for an LM to effectively capture subtle variations in academic performance that may not be discernible solely from their cognitive trajectory data. The non-cognitive data was sourced from a smartphone-based application. The privacy-preserving app triggered contextually tailored, study-specific daily questions, following rules stipulated by researchers. Participants' anonymized responses were securely aggregated on cloud-based servers for subsequent analysis.
%\vspace{-2.00mm}

\subsection{Data Pre-processing}

The cognitive data, represented numerically, and the background and non-cognitive data, expressed in natural language, required alignment. Thus, we \textbf{verbalized} the cognitive data, converting numerical scores into natural language descriptions. For instance, scores of 1/1, 3/3, and 0.8/1 in Homework 1, Lab 1, and Quiz 1, respectively, were verbalized as ``\textit{The scores are 1 out of 1 in Homework\_ 1, 3 out of 3 in Lab\_1, and 0.8 out of 1 in Quiz\_1.}'' This verbalization enables the integration of cognitive data with the text-based background and non-cognitive data. Specifically, static background text data was prepended to the longitudinal cognitive and non-cognitive data to form the input text sequence $X$.

Given the large 78-dimensional feature space and the limitations of the input context window sizes of the LMs used (e.g., the encoder-decoder LM FLAN-T5~\cite{chung_scaling_2022} used in this research can only accommodate 512 tokens), we selected a subset of features to keep the number of tokens within 512. This selection includes 5-dimensional distal background factors (class standing, major, gender, race, and family yearly income), 10-dimensional cognitive factors spanning over the first 4 weeks of the semester (first 2 Diaries, 3 Labs, 2 Quizzes, and 3 Homework Assignments), and 3-dimensional experiential non-cognitive factors (i.e., repeated measures of students' three types of engagement factors—behavioral, emotional, and cognitive). We used responses from Monday, Thursday, and Saturday. 

%Table \ref{tab:experiential} shows the questions, rules, and response choices to collect measures on the non-cognitive attributes. We used responses from Monday, Thursday, and Saturday. 

The output text sequence $Y$ reflects the student's end-of-semester final letter grade, categorized into four performance groups. Finally, the input and output data sequences were combined to create a language dataset. To assess how early in the semester LMs can accurately forecast performance, we created datasets based on 2-week, 3-week, and 4-week-long input sequences, adjusting the number of cognitive features accordingly.

%\vspace{-2.00mm}

\subsection{Data Enrichment}

Our data enrichment method is designed to enhance LM adaptation, comprising task instructions, contextual cues, a strategy for handling missing values, and dataset augmentation.

\vspace{1.00mm}
\noindent \textbf{Task Instructions and Contextual Cues:}
We incorporated a task instruction at the start of each input data sequence $X$, which reads: ``\textit{Forecast the student's end-of-semester academic performance.}'' This instruction is crucial in adapting the LM for fine-tuning based on instructional cues~\cite{wei_finetuned_2022}. Additionally, \textbf{contextual messages} were incorporated for clarity, such as ``\textit{Background information:}'' at the start of background information, and temporal cues like ``\textit{In week [WEEK\_NUMBER]}'' and ``\textit{On [NAME\_OF\_THE\_DAY]}'' for weekly and daily data, respectively. The output sequence $Y$ is contextualized with expressions like ``\textit{At the end of the semester, the student will be [STUDENT'S\_PERFORMANCE]}''

\vspace{1.00mm}
\noindent \textbf{Replacing Missing Values with Contextually Relevant Descriptors.}
Our longitudinal experiential dataset, which includes weekly responses to three non-cognitive questions, exhibited a considerable incidence of missing values. These gaps primarily arose when participants either skipped questions or temporarily uninstalled the data-collection app. In the initial week, for instance, students omitted responses to 66\% of the non-cognitive questions. Moreover, more than 37\% of participants skipped at least one such question over a period extending beyond two weeks. To address this issue, we eschewed traditional data imputation strategies in favor of inserting a contextually relevant descriptor text, specifically \textit{``Skipped the question''}, wherever data was absent. This approach was congruent with the nature of how missing values manifested in our dataset. Our decision to refrain from standard imputation methods, like Last Observation Carried Forward (LOCF)~\cite{Liu_2016}, was informed by scenarios where entire sets of daily responses were missing, rendering methods like LOCF unsuitable.

%These gaps primarily arose when participants either skipped questions or temporarily uninstalled the data-collection app (detailed information is provided in the Supplementary Material).

\vspace{1.00mm}
\noindent \textbf{Augmenting the Language Dataset.}
To address the unbalanced distribution in the initial dataset (out of 48 instances 24 outstanding, 12 average, 6 prone-to-risk, and 6 at risk), we employed oversampling with random sampling techniques~\cite{haixiang_2017} and synonym replacement for token variation~\cite{Li_2022}. This resulted in a near-balanced distribution of performance categories, reducing potential biases in LM predictions. The augmented dataset comprises 144 samples (48 outstanding, 36 average, 30 prone-to-risk, and 30 at-risk).

%Hernandez_2013

%\footnote{The dataset will be available upon paper acceptance.}

%\vspace{-2.00mm}

\section{Experiments}
\label{experiments}

This section presents a series of experiments designed to investigate four key research questions about the learning behaviors of LMs, as detailed in Section \ref{intro}. Our experimental setup involved two distinct types of pre-trained LMs: the decoder-only  LLaMA 2 (Large Language Model Meta AI)~\cite{touvron_2023_llama} and the encoder-decoder FLAN-T5~\cite{chung_scaling_2022}. These selections allowed us to compare the performance between a large-scale LM and a moderately-sized LM, specifically the 7 billion-parameter (7B) LLaMA 2 and the 770 million-parameter (770M) FLAN-T5. We fine-tuned these models across three different language datasets of varying lengths: 4-week, 3-week, and 2-week durations, aiming to assess the adaptability of LMs over diverse time frames. The performance of these adapted LMs was evaluated based on their ability to generate outputs with matching keywords corresponding to predefined performance types.

\vspace{1.00mm}
\noindent \textbf{Test Datasets.} For our testing purposes, we curated datasets by sampling approximately 30\% of instances from the augmented datasets, ensuring a balanced distribution across different classes. The rest, constituting 70\% of the data, was employed for the fine-tuning process of the LMs.

\vspace{1.00mm}
\noindent \textbf{Experimental Setup.}
For the decoder-only LM, we utilized the 7B LLaMA 2 model~\cite{touvron_2023_llama}, characterized by a maximum token limit of 4,096 in its context window. We fine-tuned this model using a parameter-efficient fine-tuning (PEFT) method QLoRA~\cite{dettmers_2023_qlora} with the following parameter settings: \texttt{lora\_r = 16, lora\_alpha = 64, lora\_dropout = 0.1, task\_type =}\allowbreak ``CAUSAL\_LM''. The model's learning rate was set to 2e-4, and the optimizer used was \texttt{paged\_adamw\_32bit}. 

In the case of the encoder-decoder LM, the 770M FLAN-T5 model~\cite{chung_scaling_2022}, a variation of the T5 model~\cite{raffel_exploring_2020}, was selected. This model has a context window capping at 512 tokens. We fine-tuned the FLAN-T5 using an AdamW optimizer~\cite{loshchilov_2018} with a learning rate set to 3e-4.

All experiments were conducted with a batch size of 4, spanning 50 epochs. This batch size was chosen in consideration of the memory limitations encountered during the fine-tuning phase. We utilized Tesla V100 (32GB RAM) and A40 (48GB RAM) GPUs for distributed training. The fine-tuning process for each model was completed in under an hour.

%\vspace{-2.00mm}

\subsection{Results}

\begin{figure*}[ht!]
    \begin{minipage}{0.5\textwidth}
        \centering
        \includegraphics[width=\linewidth]{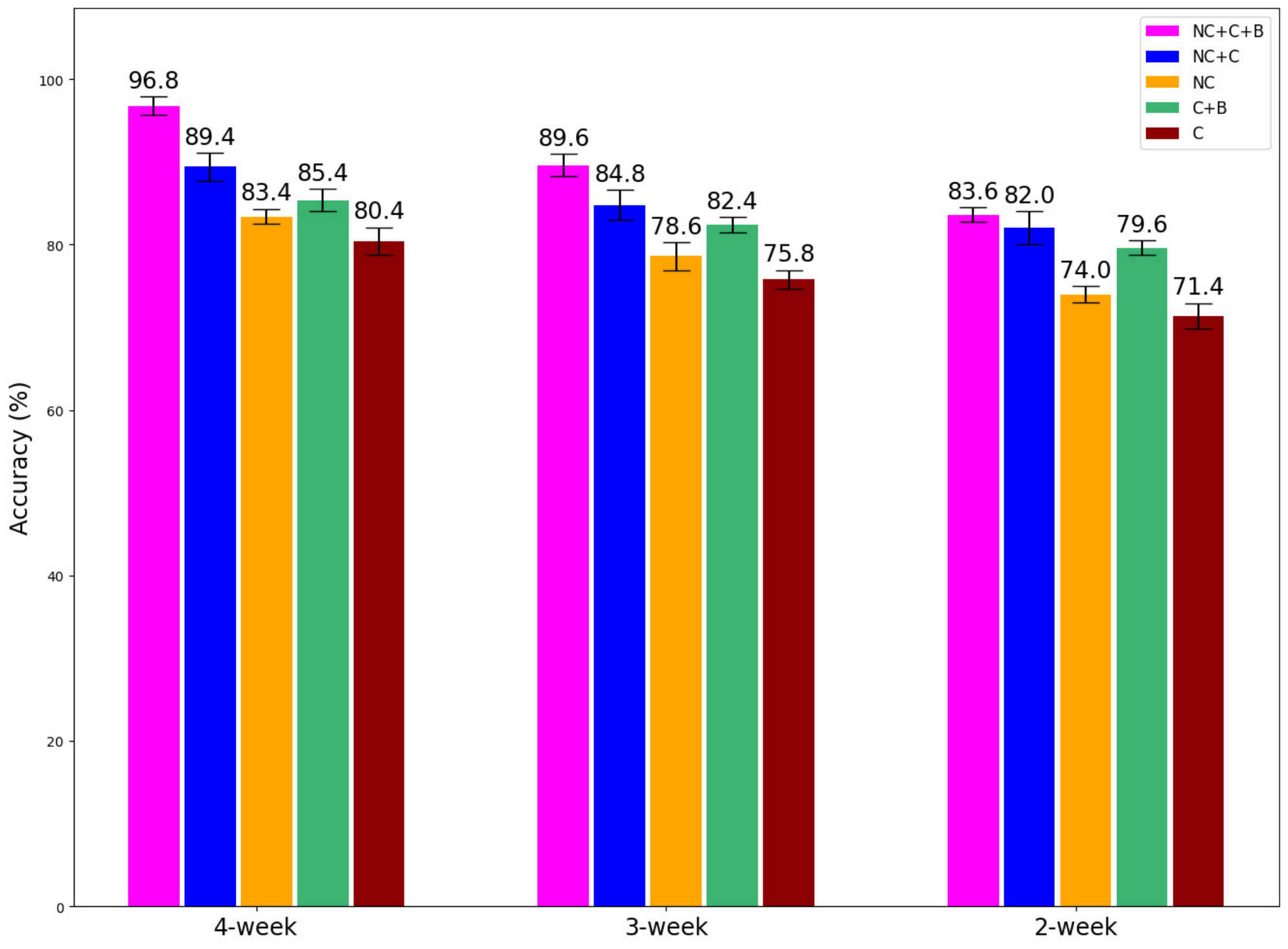}
        \caption*{(a) LLaMA}
        \label{fig:Acccuracy-LLaMA2-7B}
    \end{minipage}\hfill
    \begin{minipage}{0.5\textwidth}
        \centering
        \includegraphics[width=\linewidth]{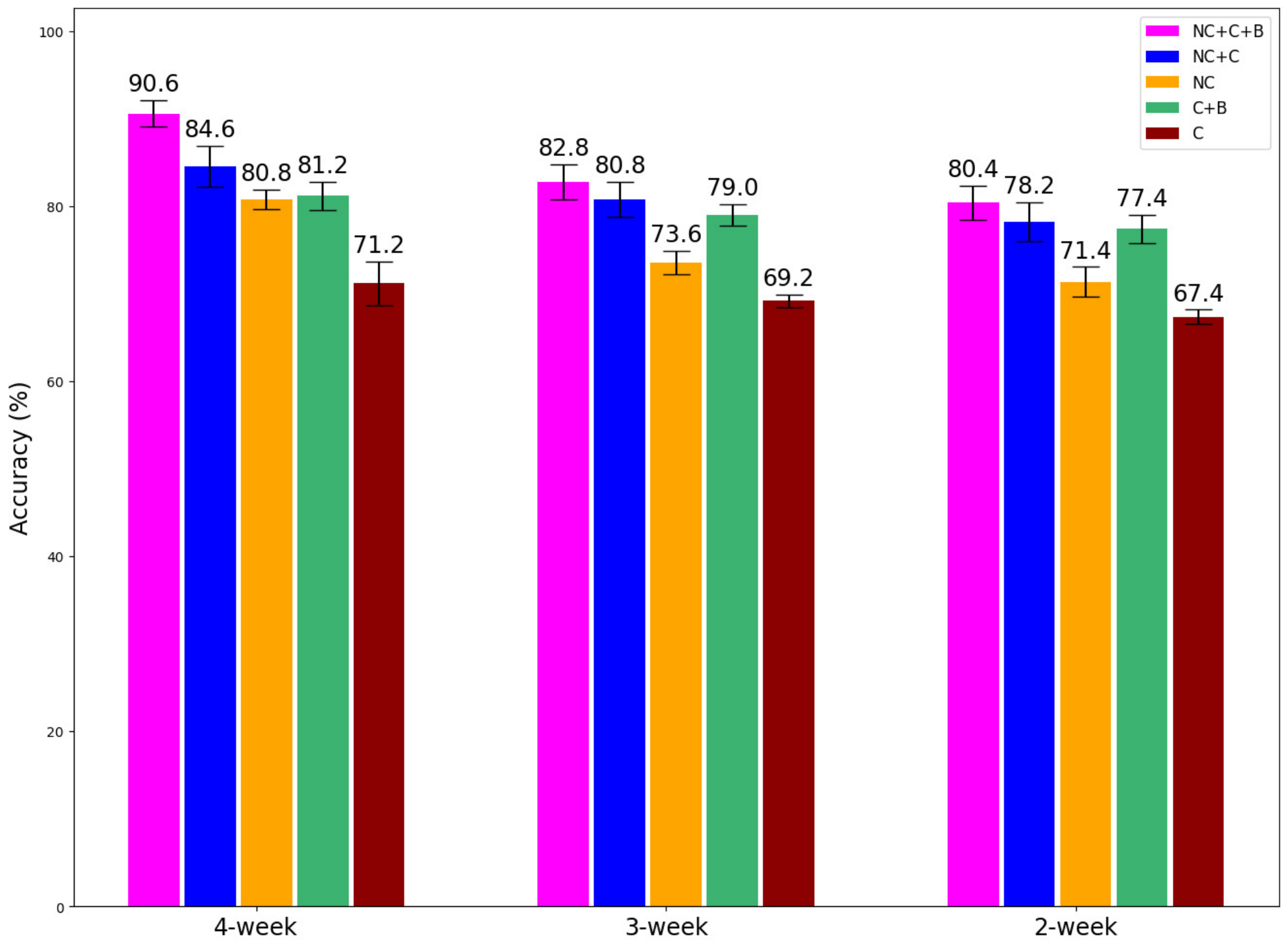}
        \caption*{(b) FLAN-T5}
        \label{fig:Acccuracy-FLAN-T5-770M}
    \end{minipage}
    \caption{[RQ1 \& RQ2]: Evaluation of two types of LMs that are fine-tuned with various combinations of experiential and non-experiential modalities of data using the 4-week, 3-week, and 2-week datasets. Each model is evaluated 5 times and the average and standard deviation are reported. \textit{Legends: NC=Non-Cognitive, C=Cognitive, B=Background}.}
    \label{fig:Acccuracy-LLaMA2-7B-FLAN-T5-770M}
    \vspace{-2.00mm}

\end{figure*}

\noindent \textbf{[RQ1]: \textit{To what extent can LMs accurately forecast outcomes based solely on longitudinal experiential data?}}

To explore RQ1 and RQ2, we examined the forecasting abilities of two types of LMs (LLaMA 2 and FLAN-T5) using both the three modalities of longitudinal experiential data. Our approach involved fine-tuning the LMs across five distinct combinations of the three data modalities and employing language datasets of varying durations (4-, 3-, and 2-week).

Our findings, illustrated in Figure \ref{fig:Acccuracy-LLaMA2-7B-FLAN-T5-770M}, reveal a notable trend: models fine-tuned exclusively with non-cognitive data consistently outperformed those trained solely on cognitive-only data. This outcome underscores the LMs' remarkable ability to extract meaningful insights into students' academic progress by analyzing their experiential data. What makes this even more significant is the limited feature set used for adaptation – only three out of twenty-eight available features related to motivation and engagement were employed.

This performance was particularly striking in the case of the 2-week language datasets. Despite grappling with a 66\% rate of missing values in the first week's non-cognitive data, the LMs demonstrated impressive forecasting accuracy for end-of-semester cognitive performance. The accuracy rates exceeded 70\% on average, with LLaMA achieving 74.0\% and FLAN-T5 reaching 71.4\%. In contrast, the cognitive-only models lagged slightly behind, with LLaMA at 71.4\% and FLAN-T5 at 67.4\%.
\vspace{2.00mm}

\noindent \textbf{[RQ2]: \textit{How effectively do LMs capture and leverage correlations across the non-cognitive, cognitive, and background modalities within academic experiential data for precise early forecasting?}}

Our initial hypothesis posited that the key to accurate forecasting lies in the LMs’ ability to learn and integrate correlations across data modalities. The results, as depicted in Figure \ref{fig:Acccuracy-LLaMA2-7B-FLAN-T5-770M}, affirm this hypothesis. We observed that both LLaMA and FLAN-T5 LMs achieved their peak performance when fine-tuned with a combination of three modalities, highlighting the importance of a multi-modal approach.

In terms of comparative performance, LLaMA consistently outshone FLAN-T5. A distinguished milestone was reached as early as week 2, with the LLaMA model achieving an impressive 83.6\% accuracy in its forecasts, while FLAN-T5 followed closely with 80.4\% accuracy. The peak of this performance was observed with the week-4 data, where LLaMA reached a remarkable 96.8\% accuracy. This high level of accuracy achieved by LLaMA underscores the efficacy of our approach in leveraging the full spectrum of data modalities for early and precise academic performance forecasting.

\begin{figure*}[htb!]
    \begin{minipage}{0.5\textwidth}
        \centering
        \includegraphics[width=\linewidth]{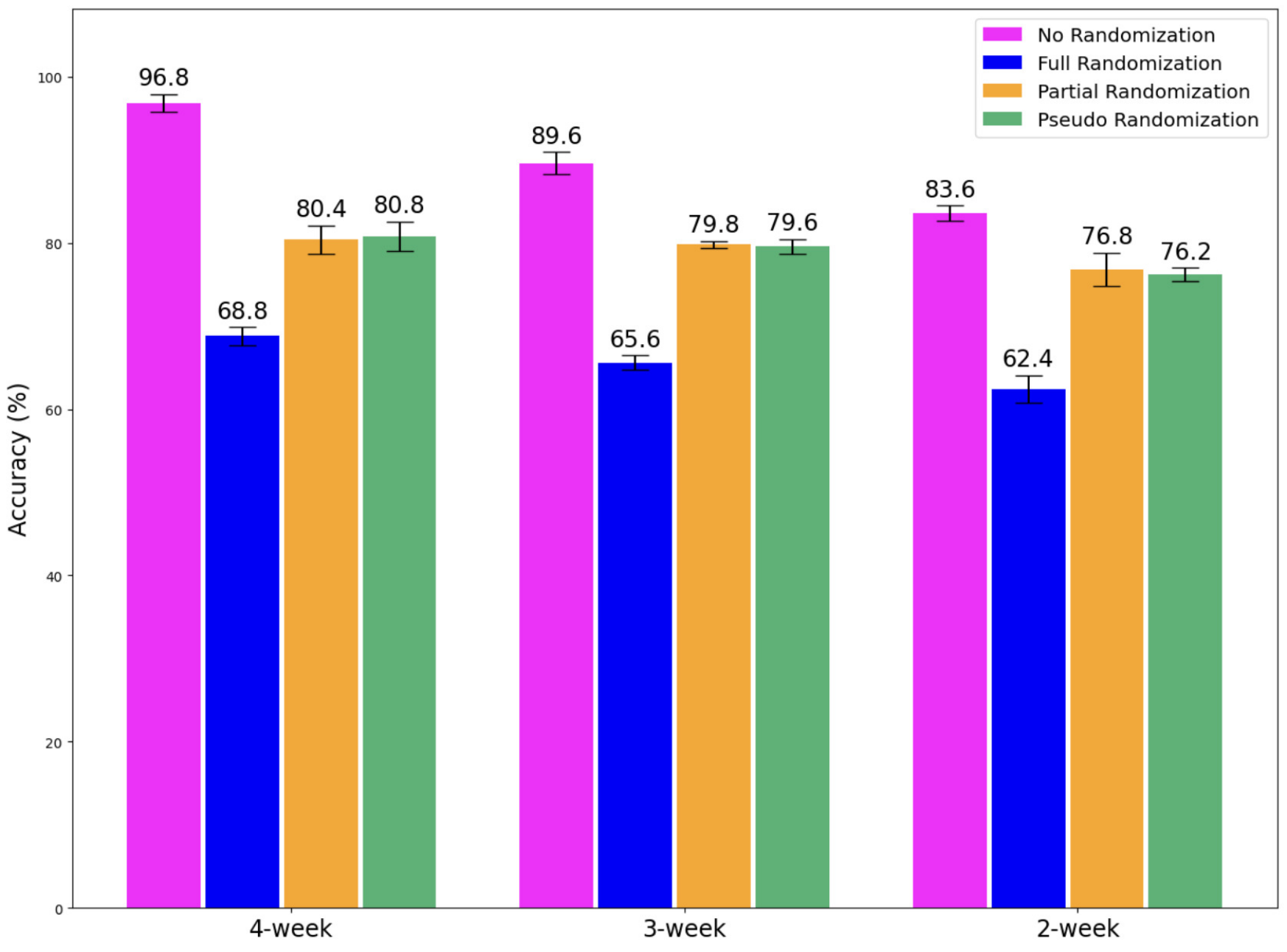}
        \caption*{(a) LLaMA}
        \label{fig:Temporal-LLaMA2-7B}
    \end{minipage}\hfill
    \begin{minipage}{0.5\textwidth}
        \centering
        \includegraphics[width=\linewidth]{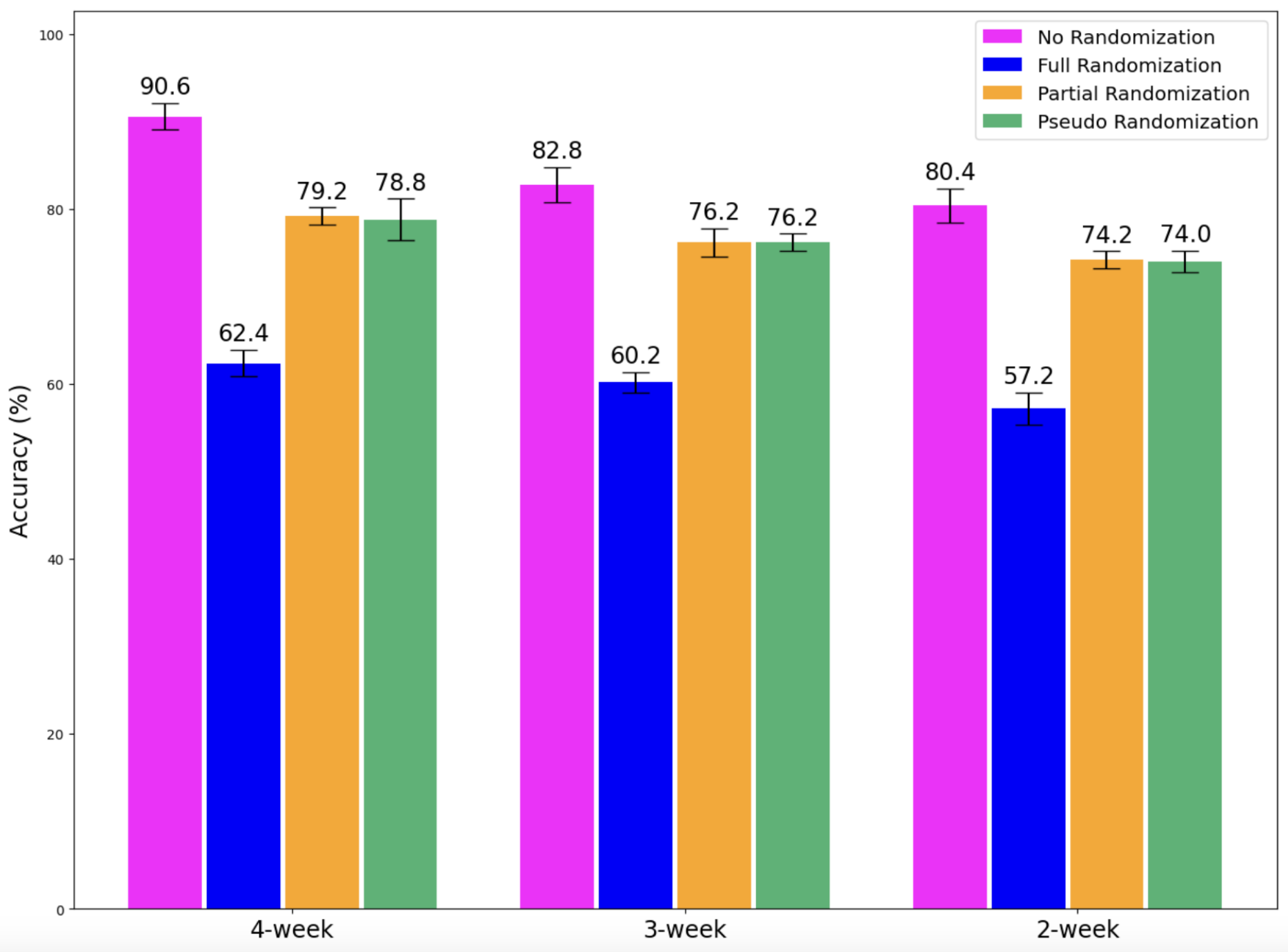}
        \caption*{(b) FLAN-T5}
        \label{fig:Temporal-FLAN-T5-770M}
    \end{minipage}
    \caption{[RQ3]: Performance of the LMs (fine-tuned with both modalities) is compared with LMs fine-tuned with randomized measures of three experiential modalities.}
    \label{fig:Temporal-LM}
    \vspace{-6.00mm}
\end{figure*}

\vspace{2.00mm}
\noindent \textbf{[RQ3]: \textit{What is the extent of LMs' ability to interpret and use temporal variations within the dataset for forecasting purposes?}}

The impressive accuracy demonstrated by the LMs (refer to Figure \ref{fig:Acccuracy-LLaMA2-7B-FLAN-T5-770M}) raises an intriguing question: Do these models genuinely grasp and utilize the temporal dynamics in the dataset, or are they merely capturing broad statistical patterns? This inquiry is crucial, considering that our dataset was enriched with contextual temporal markers to signify the chronological progression of weeks (e.g., ``In week 1'') and days within a week (e.g., ``On Monday''). We anticipated that these temporal cues would be instrumental for the LMs in discerning and leveraging the nuanced variations in the time-series data.

%To probe deeper into this, we have devised a series of three experiments. Each experiment employs a distinct version of the dataset, where we systematically increase the level of temporal randomization. Moreover, in the first two experiments, we intentionally omit the temporal markers to assess the models' ability to perceive temporal sequences without explicit cues. These experiments are designed to unravel the extent to which the LMs are sensitive to the temporal ordering of data, whether they can still perform effectively when this order is disrupted, and crucially, their proficiency in harnessing explicit temporal cues when available.

To probe deeper into this aspect, we have crafted a trio of experiments, each utilizing a different iteration of the dataset that gradually enhances the degree of temporal randomization. Moreover, in the first two experiments, we intentionally omit the temporal markers to assess the models' ability to perceive temporal sequences without explicit cues. These experiments are designed to unravel the extent to which the LMs are sensitive to the temporal ordering of data, whether they can still perform effectively when this order is disrupted, and crucially, their proficiency in harnessing explicit temporal cues when available.

\begin{figure*}[!htb]
    \centering
    \begin{minipage}{0.33\textwidth}
        \centering
        \includegraphics[width=\linewidth]{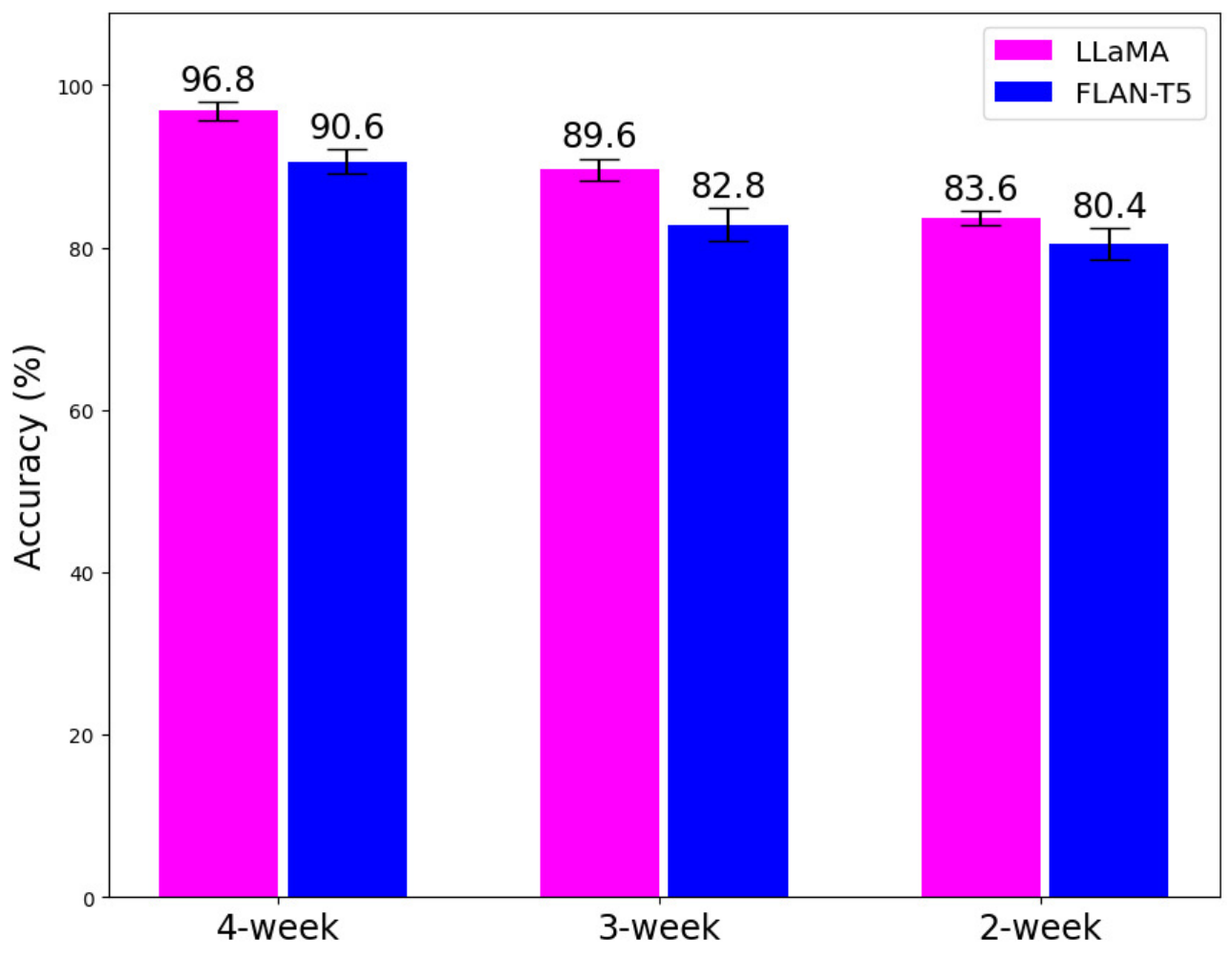}
        \caption*{(a) ``Skipped the question''}
        \label{fig:Missing-Value-Skipped}
    \end{minipage}\hfill
    \begin{minipage}{0.33\textwidth}
        \centering
        \includegraphics[width=\linewidth]{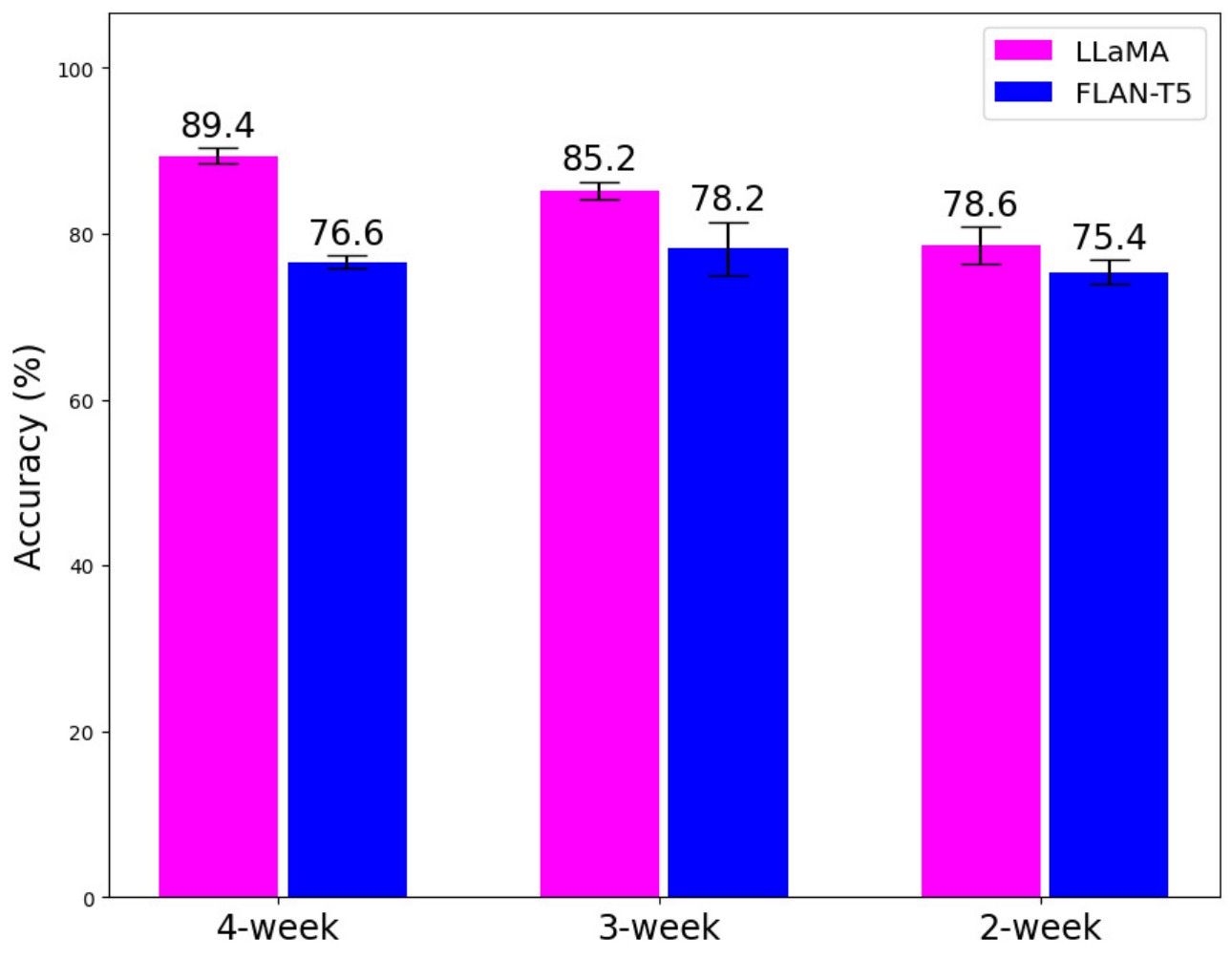}
        \caption*{(b) ``N/A''}
        \label{fig:Missing-Value-NA}
    \end{minipage}\hfill
    \begin{minipage}{0.33\textwidth}
        \centering
        \includegraphics[width=\linewidth]{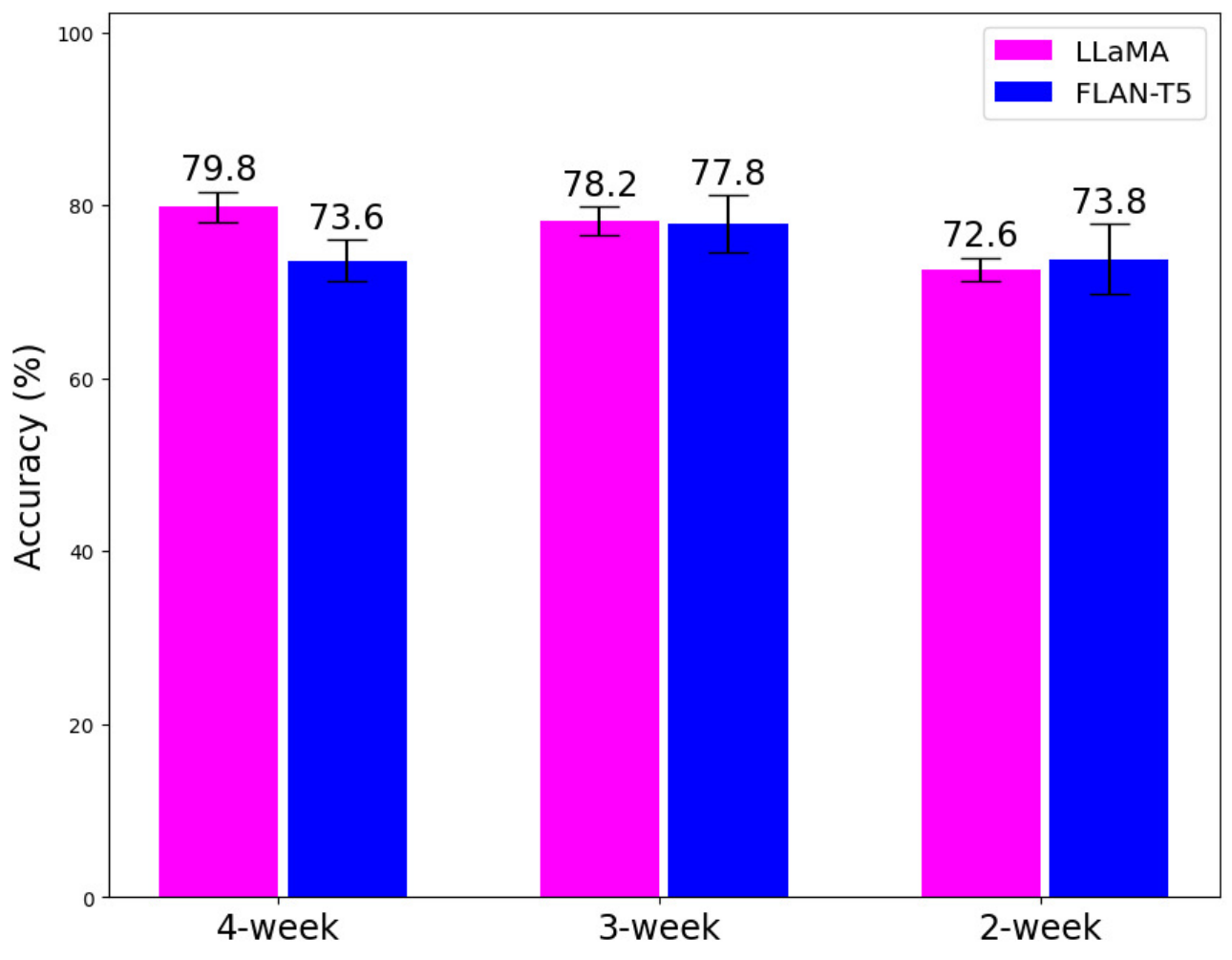}
        \caption*{(c) ``Hello, World!''}
        \label{fig:Missing-Value-Hello-World}
    \end{minipage}
   \vspace{-1.00mm}
    \caption{\textbf{[RQ4]}: Investigation of the impact of the contextually relevant descriptor of missing values.}
    \label{fig:Missing-Value}
    \vspace{-5.00mm}
\end{figure*}

%\vspace{-2.00mm}
\begin{itemize}
\vspace{-1.00mm}
\item Experiment RQ3(a) [\textbf{Full randomization}]: This experiment introduces complete randomization by shuffling both the order of weeks and days within each week. Additionally, we omit temporal markers such as ``In Week [WEEK\_COUNT]'' to eliminate any explicit chronological cues.

\item Experiment RQ3(b) [\textbf{Partial randomization}]: Here, we only randomize the sequence of weeks while maintaining the day-to-day order within each week. Similar to RQ3(a), we remove weekly temporal tags to obscure the original temporal sequence.

\item Experiment RQ3(c) [\textbf{Pseudo randomization}]: This setup mirrors the partial randomization, except that we retain the weekly tags. Despite the shuffled order of weeks, these tags could potentially aid the models in discerning the chronological sequence, hence the term ``pseudo randomization.''

\end{itemize}

For both LLaMA and FLAN-T5 models, we evaluated their performance on these three randomized datasets against a baseline trained on the original, chronologically ordered dataset (no randomization). The results, depicted in Figure \ref{fig:Temporal-LM}, are revealing. With full randomization (Experiment RQ3(a)), there's a significant performance drop (ranging between 21\% to 28\% across different time frames) compared to the baseline. This decline highlights the impact of temporal disarray on the models' forecasting abilities.

In the partial randomization scenario (Experiment RQ3(b)), the performance decrement is less severe, under 10\% for both models across 2-week and 3-week datasets. However, for the 4-week dataset, the decline is more pronounced, reaching up to 16.4\% for LLaMA and 11.4\% for FLAN-T5. This trend suggests that retaining daily sequences within each week mitigates the negative impact of disrupted weekly sequences, though not completely.

%Unsurprisingly again, in the partial randomization experiment, the models' accuracies did not decline significantly from that of the full randomization case because though weekly orders were randomized in partial randomization, the daily orders in each week were retained. Moreover, since we removed the semantically useful contextual tags informing weekly orders, the models could not accurately discern the weekly temporal patterns, which explains the observed drop in performance from the no randomization case.

Interestingly, in the pseudo randomization experiment (RQ3(c)), despite preserving the weekly tags, there was no significant improvement in model accuracy over the partially randomized scenario. In some cases, such as with the LLaMA model, performance even slightly declined (by $0.2\%$ in 3 weeks and $0.6\%$ in 2 weeks). Thus, even with its comparatively larger base of general knowledge and better reasoning ability, the 7B LLaMA does not seem to pick up the explicitly encoded temporal signals. This outcome suggests that \textbf{while the models are capable of learning statistical patterns from time-varying measures, they may struggle to fully comprehend and utilize explicit temporal cues encoded in the data}.

%\noindent \textbf{[RQ4]: \textit{How do LMs utilize contextually relevant descriptors to compensate for missing values in the data?}}
\vspace{2.00mm}
\noindent \textbf{[RQ4]: \textit{How can we effectively address missingness in experiential datasets by leveraging pre-trained LMs?}}

A significant challenge in our dataset is the presence of numerous missing values in its experiential dimension. We explore how the general knowledge embedded in LMs can be harnessed to address this issue. Prior research has demonstrated that LMs are adept at managing missing values, often by substituting them with a generic descriptor like ``N/A''~\cite{gruver_2023_llm}. Our focus here is on evaluating whether contextually nuanced descriptors for missing data enhance the performance of LMs in comparison to standard descriptors, and to what extent incorrect descriptors influence model accuracy.

Three experiments were designed to investigate this:

\begin{itemize}

\item Experiment RQ4(a) [\textbf{Replacement with ``Skipped the question''}]: This experiment (Figure  \ref{fig:Missing-Value}(a)) involved substituting missing values with the phrase ``Skipped the question'', a contextually relevant descriptor. 

\item Experiment RQ4(b) [\textbf{Replacement with ``N/A''}]: In this setup (Figure  \ref{fig:Missing-Value}(b)), missing values were replaced with the more generic but still contextually correct ``N/A''.

\item Experiment RQ4(c) [\textbf{Replacement with ``Hello, World!''}]: This experiment (Figure  \ref{fig:Missing-Value}(c)) used ``Hello, World!'' as an intentionally contextually incorrect descriptors for missing values.

%\item Experiment RQ4(a) [\textbf{Replacement with ``Skipped the question''}]: This experiment (Figure \ref{fig:Missing-Value-Skipped}) involved substituting missing values with the phrase ``Skipped the question'', a contextually relevant descriptor. Figure  \ref{fig:Missing-Value}
%
%\item Experiment RQ4(b) [\textbf{Replacement with ``N/A''}]: In this setup (Figure \ref{fig:Missing-Value-NA}), missing values were replaced with the more generic but still contextually correct ``N/A''.
%
%\item Experiment RQ4(c) [\textbf{Replacement with ``Hello, World!''}]: This experiment (Figure \ref{fig:Missing-Value-Hello-World}) used ``Hello, World!'' as an intentionally contextually incorrect descriptors for missing values.

\end{itemize}

The results reveal some intriguing patterns. The first experiment (RQ4(a)) using ``Skipped the question'' led to the highest performance in both LLaMA and FLAN-T5 models, underscoring the value of contextually rich descriptors. Surprisingly, even though ``N/A'' is contextually correct, its use resulted in a marked performance drop (RQ4(b)). For instance, accuracy for the 4-week LLaMA model decreased by 7.4\%, and for the FLAN-T5 model, the decline was even more significant at 14\%. This suggests a sensitivity of these models to the specific wording used in missing value descriptors.
%\vspace{-2.00mm}

Furthermore, in RQ4(b) and RQ4(c), despite using an incorrect descriptor, the performance decrease was minor: less than 3\% for FLAN-T5 and less than 10\% for LLaMA. This suggests that LMs maintain robustness even with missing data, regardless of the descriptor’s contextual accuracy.

%\vspace{-3.00mm}
%\section{Interpretability Framework}
\section{Conclusion}
\label{conclusion}

This research explores the capabilities of pre-trained LMs for early forecasting of academic trajectories in STEM students using longitudinal experiential data. Leveraging a novel dataset that encompasses non-cognitive, cognitive, and background factors, we fine-tuned LMs through an innovative data enrichment process that addressed missing values, augmented textual sequences, and incorporated task-specific instructions and contextual indicators. Our findings reveal that while LMs effectively integrate multiple data modalities and demonstrate robustness in handling incomplete data, they primarily rely on high-level statistical patterns, lacking deeper semantic understanding of temporal dynamics. Additionally, their ability to interpret explicit temporal information remains limited. Moving forward, expanding the dataset will be critical to improving fine-tuning and gaining deeper insights, particularly into non-cognitive features.

\section*{Acknowledgments}
This research was supported by grants from the U.S. National Science Foundation (NSF DUE 2142558), the U.S. National Institutes of Health (NIH NIGMS P20GM130461 and NIH NIAAA R21AA029231), and the Rural Drug Addiction Research Center at the University of Nebraska-Lincoln. 

%The content is solely the responsibility of the authors and does not necessarily represent the official views of the National Science Foundation, National Institutes of Health, or the University of Nebraska.

% BibTeX users should specify bibliography style 
%\vspace{-2.00mm}

\bibliographystyle{IEEEtran}
\bibliography{references}

\end{document}